\documentclass[10pt,twocolumn,letterpaper]{article}

\usepackage[colorlinks]{hyperref}
\usepackage{cvpr}
\usepackage{times}
\usepackage{eso-pic}
\usepackage{xspace}
\usepackage{epsfig}
\usepackage{graphicx}
\usepackage{amsmath}
\usepackage{amssymb}
\usepackage{float}
\usepackage{multirow}
\usepackage{rotating}
\usepackage{balance}
\usepackage{wrapfig}
\usepackage{enumerate}
\usepackage{caption}

\DeclareMathOperator*{\argmax}{arg\,max}

\def\cvprPaperID{3417} 

\cvprfinalcopy 
\ifcvprfinal\fi

\begin{document}

\title{Commonly Uncommon:\\Semantic Sparsity in Situation Recognition}

\author{Mark Yatskar$^1$,  Vicente Ordonez$^{2,3}$, Luke Zettlemoyer$^1$, Ali Farhadi$^{1,2}$ \\
        $^1$Computer Science \& Engineering, University of Washington, Seattle, WA\\
        $^2$Allen Institute for Artificial Intelligence (AI2), Seattle, WA\\
        $^3$Department of Computer Science, University of Virginia, Charlottesville, VA. \\
	    {\tt [my89, lsz, ali]@cs.washington.edu, vicente@cs.virginia.edu}
  }

\maketitle

\begin{abstract}
Semantic sparsity is a common challenge in structured visual classification problems; when the output space is complex, the vast majority of the possible predictions are rarely, if ever, seen in the training set. 
This paper studies semantic sparsity in situation recognition, the task of producing structured summaries of what is happening in images, including activities, objects and the roles objects play within the activity. 
For this problem, we find empirically that most object-role combinations are rare, and current state-of-the-art models significantly underperform in this sparse data regime.
We avoid many such errors by (1) introducing a novel tensor composition function that learns to share examples across role-noun combinations and (2) semantically augmenting our training data with automatically gathered examples of rarely observed outputs using web data.
When integrated within a complete CRF-based structured prediction model, the tensor-based approach outperforms existing state of the art by a relative improvement of 2.11\%  and 4.40\% on top-5 verb and noun-role accuracy, respectively.
Adding 5 million images with our semantic augmentation techniques gives further relative improvements of 6.23\% and 9.57\% on top-5 verb and noun-role accuracy.
\end{abstract}

\section{Introduction}
Many visual classification problems, such as image captioning \cite{mscoco}, visual question answering \cite{vqa}, referring expressions \cite{referit}, and situation recognition \cite{yatskar2016} have structured, semantically interpretable output spaces. 
In contrast to classification tasks such as ImageNet~\cite{imagenet}, these problems typically suffer from \textit{semantic sparsity}; there is a combinatorial number of possible outputs, no dataset can cover them all, and performance of existing models degrades significantly when evaluated on rare or unseen inputs~\cite{atzmon2016learning,zhou2015simple,devlin2015exploring,yatskar2016}. In this paper, we consider situation recognition, a prototypical structured classification problem with significant semantic sparsity, and develop new models and semantic data augmentation techniques that significantly improve performance by better modeling the underlying semantic structure of the task.


\begin{figure}[t]
\centering
\hspace*{-10pt}
\includegraphics[width=0.48\textwidth]{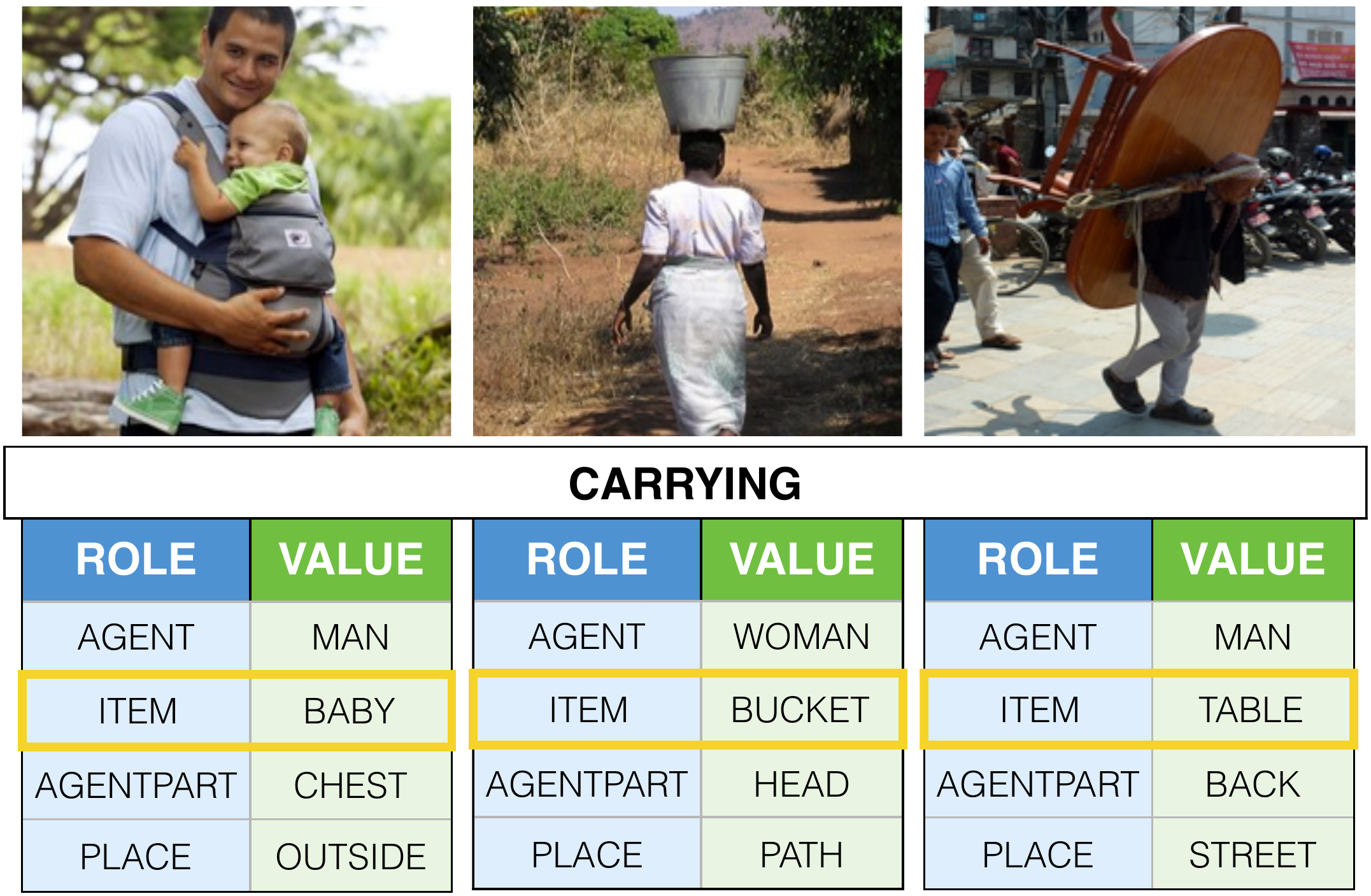}
\caption{\small Three situations involving \texttt{carrying}, with semantic roles \texttt{agent}, the carrier, \texttt{item}, the carried, \texttt{agentpart}, the part of the agent carrying, and \texttt{place}, where the situation is happening. For carrying, there are many possible carry-able objects (nouns that can fill the \texttt{item} role), which is an example of semantic sparsity that holds for many roles in situation recognition.}
\label{fig:leading}
\end{figure}

\begin{figure}[t]
\centering
\includegraphics[width=0.5\textwidth]{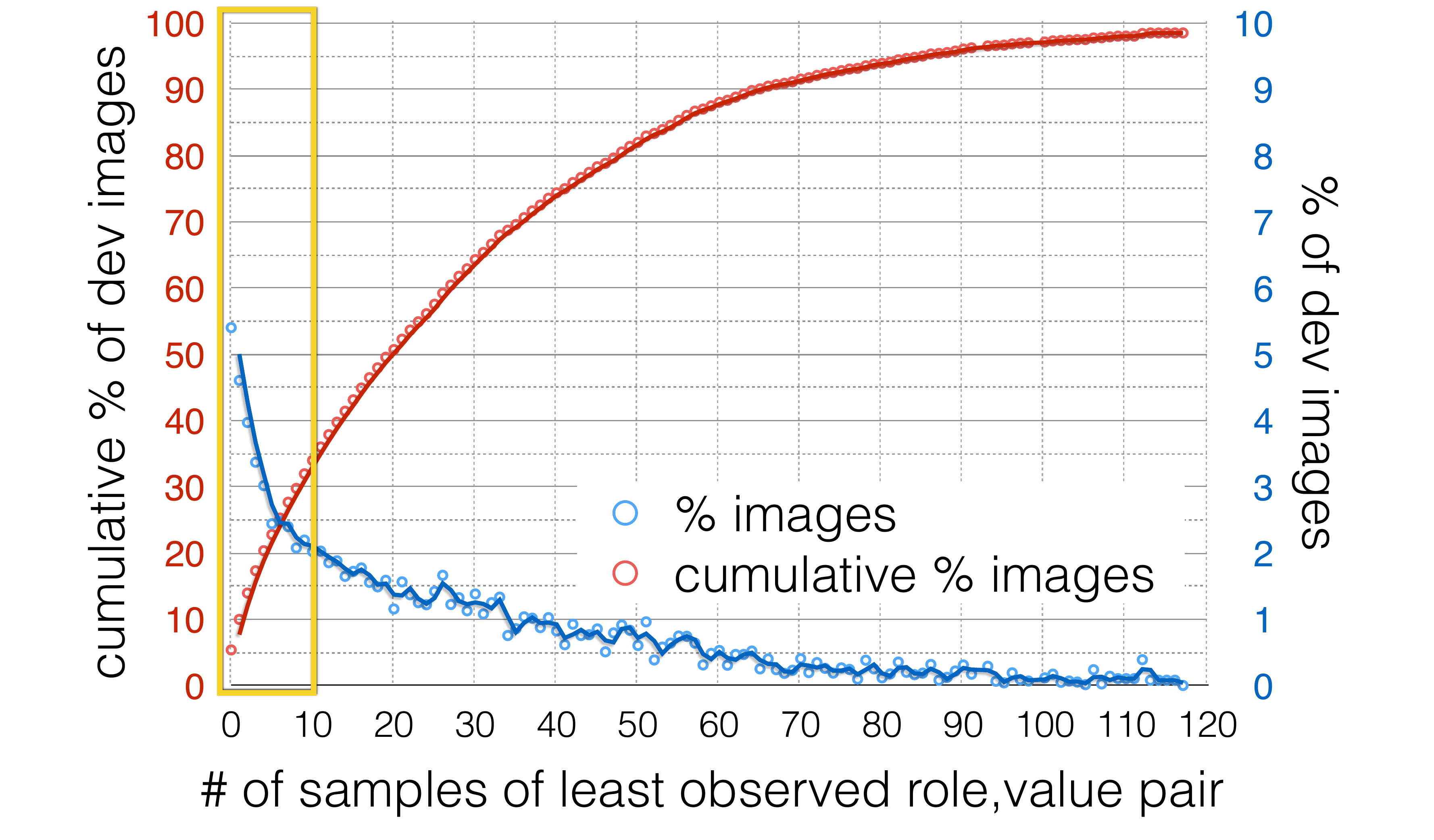}
\caption{\small The percentage of images in the imSitu development set as a function of the total number of training examples for the least frequent role-noun pair in each situation.
Uncommon target outputs, those observed fewer than 10 times in training (yellow box), are common, constituting 35\% of all required predictions. Such semantic sparsity is a central challenge for situation recognition. }
\label{fig:freq}
\end{figure}
\begin{figure}[t]
\centering
\includegraphics[width=0.5\textwidth]{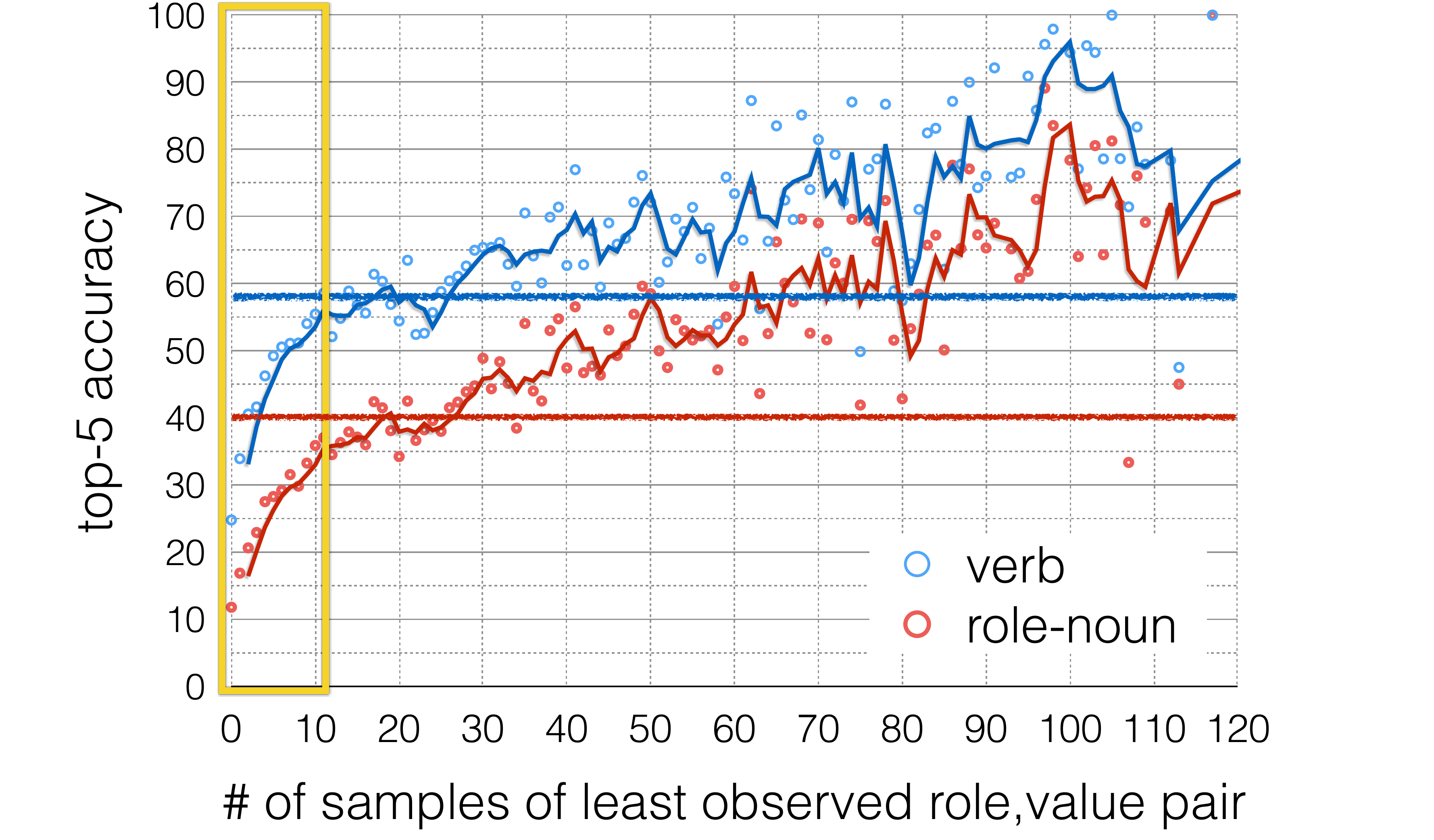}
\caption{\small Verb and role-noun prediction accuracy of a baseline CRF \cite{yatskar2016} on the imSitu dev set as a function of the frequency of the least observed role-noun pair in the training set. Solid horizontal lines represent average performance across the whole imSitu dev set, irrespective of frequency. As even one target output becomes uncommon (highlighted in yellow box), accuracy decreases. }
\label{fig:oov2}
\end{figure}

Situation recognition \cite{yatskar2016} is the task of producing structured summaries of what is happening in images, including activities, objects and the roles those objects play within the activity.
This problem can be challenging because many activities, such as \texttt{carrying}, have very open ended semantic roles, such as \texttt{item}, the thing being carried (see Figure~\ref{fig:leading}); nearly any object can be carried and the training data will never contain all possibilities. This is a prototypical instance of semantic sparsity: rare outputs constitute a large portion of required predictions (35\% in the imSitu dataset~\cite{yatskar2016}, see Figure \ref{fig:freq}), and current state-of-the-art performance for situation recognition drops significantly when even one participating object has few samples for it's role (see Figure \ref{fig:oov2}). 
We propose to address this challenge in two ways by (1) building models that more effectively share examples of objects between different roles and (2) semantically augmenting our training set to fill in rarely represented noun-role combinations.



We introduce a new compositional Conditional Random Field formulation (CRF) to reduce the effects of semantic sparsity by encouraging sharing between nouns in different roles. 
Like previous work \cite{yatskar2016}, we use a deep neural network to directly predict factors in the CRF.
In such models, required factors for the CRF are predicted using a global image representation through a linear regression unique to each factor.  
In contrast, we propose a novel tensor composition function that uses low dimensional representations of nouns and roles, and shares weights across all roles and nouns to score combinations.
Our model is compositional, independent representations of nouns and roles are combined to predict factors, and allows for a globally shared representation of nouns across the entire CRF.

This model is trained with a new form of semantic data augmentation, to provide extra training samples for rarely observed noun-role combinations. 
We show that it is possible to generate short search queries that correspond to partial situations (i.e. ``man carrying baby'' or ``carrying on back'' for the situations in Figure \ref{fig:leading}) which can be used for web image retrieval.
Such noisy data can then be incorporated in pre-training by optimizing marginal likelihood, effectively performing a soft clustering of values for unlabeled aspects of situations.
This data also supports, as we will show, self training where model predictions are used to prune the set of images before training the final predictor. 

Experiments on the imSitu dataset \cite{yatskar2016} demonstrate that our new compositional CRF and semantic augmentation techniques reduce the effects of semantic sparsity, with  strong gains for relatively rare configurations. 
We show that each contribution helps significantly, and that the combined approach improves performance relative to a strong CRF baseline by 6.23\% and 9.57\% on top-5 verb and noun-role accuracy, respectively. On uncommon predictions, our methods provide a relative improvement of 8.76\% on average across all measures. Together, these experiments demonstrate the benefits of effectively targeting semantic sparsity in structured classification tasks.



\section{Background}
\label{sec:task}
\paragraph{Situation Recognition}
Situation recognition has been recently proposed to model events within images \cite{gupta_vsrl, vsrl2,yang_vsrl,yatskar2016}, in order to answer questions beyond just ``What activity is happening?'' such as ``Who is doing it?'', ``What are they doing it to?'', ``What are they doing it with?''.  
In general, formulations build on semantic role labelling \cite{srl}, a problem in natural language processing where verbs are automatically paired with their arguments in a sentence (for example, see \cite{dasthesis}). 
Each semantic role corresponds to a question about an event, (for example, in the first image of Figure \ref{fig:leading}, the semantic role \texttt{agent} corresponds to ``who is doing the carrying?'' and \texttt{agentpart} corresponds to ``how is the item being carried?''). 

We study situation recognition in imSitu \cite{yatskar2016}, a large-scale dataset of human annotated situations containing over 500 activities, 1,700 roles, 11,000 nouns, 125,000 images. 
imSitu images are collected to cover a diverse set of situations.
For example, as seen in Figure~\ref{fig:freq}, 35\% of situations annotated in the imSitu development set contain at least one rare role-noun pair. Situation recognition in imSitu is a strong test bed for evaluating methods addressing semantic sparsity: it is large scale, structured, easy to evaluate, and has a clearly measurable range of semantic sparsity across different verbs and roles.
Furthermore, as seen in Figure~\ref{fig:oov2}, semantic sparsity is a significant challenge for current situation recognition models.

\paragraph{Formal Definition}
In situation recognition, we assume a discrete sets of verbs $V$, nouns $N$, and frames $F$.
Each frame $f \in F$ is paired with a set of semantic roles $E_f$.
Every element in $V$ is mapped to exactly one $f$. 
The verb set $V$ and frame set $F$ are derived from FrameNet~\cite{framenet}, a lexicon for semantic role labeling, while the noun set $N$ is drawn from WordNet~\cite{wordnet}.
Each semantic role $e \in E_f$ is paired with a noun value $n_e \in N \cup \{\varnothing\}$, where $\varnothing$ indicates the value is either not known or does not apply.
The set of pairs of semantic roles and their values is called a realized frame, $R_f = \{(e, n_e) : e \in E_f\}$. 
Realized frames are valid only if each $e \in E_f$ is assigned exactly one noun $n_e$.

Given an image, the task is to predict a situation, $S = (v, R_f)$, specified by a verb $v \in V$ and a valid realized frame $R_f$, where $f$ refers to a frame mapped by $v$ . For example, in the first image of Figure \ref{fig:leading}, the predicted situations is 
{\small $S=$ (\texttt{carrying}, \{(\texttt{agent},\texttt{man}), (\texttt{item},\texttt{baby}), (\texttt{agentpart},\texttt{chest}), (\texttt{place},\texttt{outside})\})}.

\section{Methods}
\label{sec:model}
\begin{figure*}[t]
\centering
\includegraphics[width=.8\textwidth]{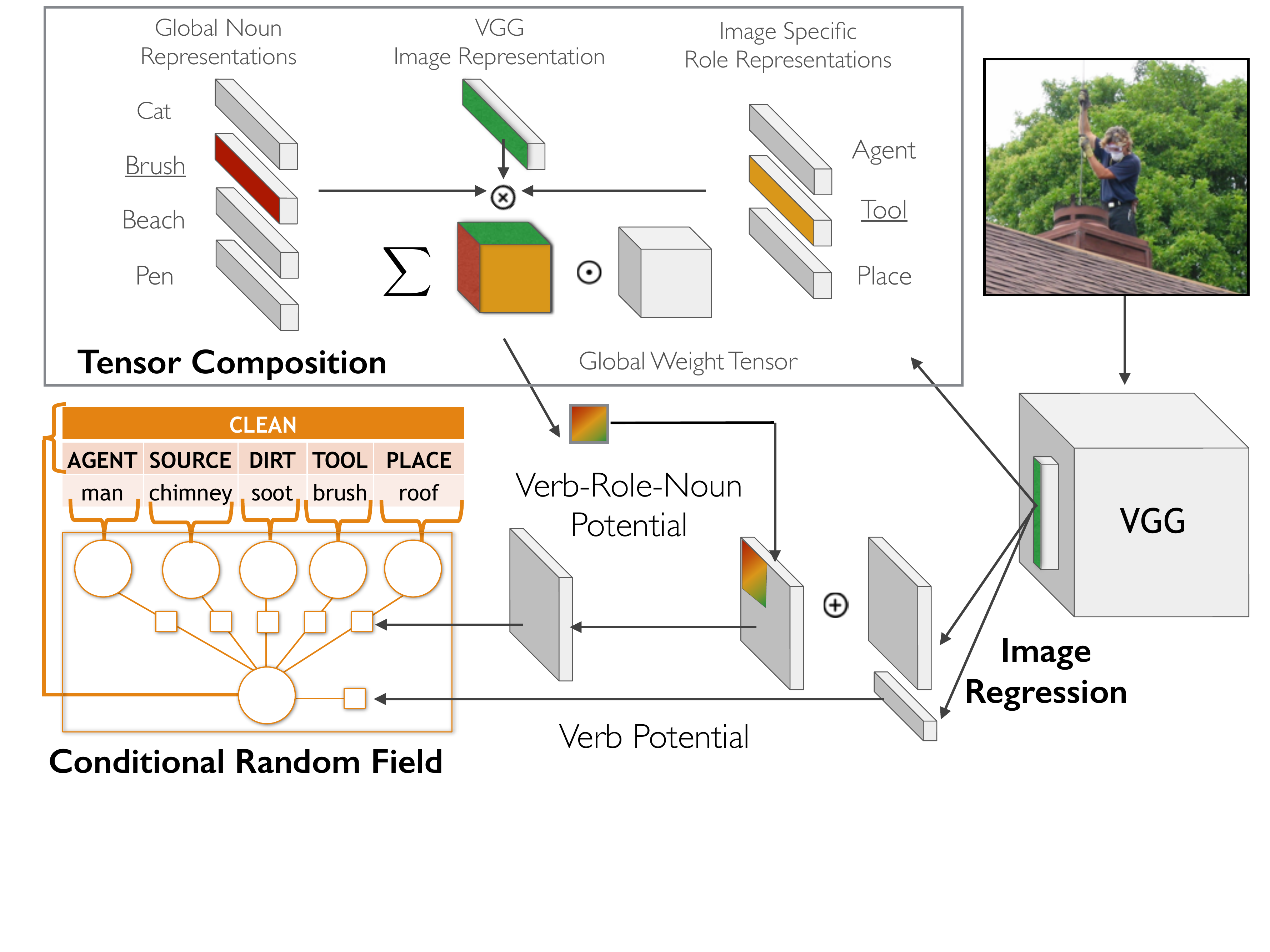}
\vspace{-.75in}
\caption{\small An overview of our compositional Conditional Random Field (CRF) for predicting situations. 
A deep neural network is used to compute potentials in a CRF.
The verb-role-noun potential is built from 
a global bank of noun representations, image specific role representations and a global image representation that are combined with a weighted tensor product. 
The model allows for sharing among the same nouns in different roles, leading to significant gains, as seen in Section \ref{sec:results}. }
\label{fig:model}
\end{figure*}

This section presents our compositional CRFs and semantic data augmentation techniques.

\subsection{Compositional Conditional Random Field}

Figure~\ref{fig:model} shows an overview of our compositional conditional random field model, which is described below.

\paragraph{Conditional Random Field} Our CRF for predicting a situation, $S = (v,R_f)$, given an image $i$, decomposes over the verb $v$ and semantic role-value pairs $(e,n_e)$ in the realized frame $R_f = \{(e, n_e) : e \in E_f\}$, similarly to previous work \cite{yatskar2016}. 
The full distribution, with potentials for verbs $\psi_v$ and semantic roles $\psi_e$ takes the form:
\begin{equation}
p(S |i; \theta) \propto \psi_v(v,i; \theta)\prod_{(e,n_e) \in R_f} \psi_e(v,e,n_e,i; \theta)
\label{eqn:crf_pot}
\end{equation}
The CRF admits efficient inference: we can enumerate all verb-semantic roles that occur and then sum all possible semantic role values that occurred in a dataset. 

Each potential in the CRF is log linear:
\begin{equation}
\psi_v(v,i; \theta) = e^{\phi_v(v,i,\theta)}
\label{eqn:pot2}
\end{equation}
\begin{equation}
\psi_e(v,e,n_e,i; \theta) = e^{\phi_e(v,e,n_e,i,\theta)}
\label{eqn:pot1}
\end{equation}
where $\phi_e$ and $\phi_v$ encode scores computed by a neural network.
To learn this model, we assume that for an image $i$ in dataset $Q$ there can, in general, be a set $A_i$ of possible ground truth situations \footnote{imSitu provides three realized frames per example image.}.
We optimize the log-likelihood of observing at least one situation $S \in A_i$:
\begin{equation}
\sum_{i \in Q} \log\Big( 1 - \prod_{S \in A_i}(1-p(S|i;\theta))\Big)
\label{eqn:crf_objective}
\vspace{-10pt}
\end{equation}


\paragraph{Compositional Tensor Potential}
In previous work, the CRF potentials (Equation \ref{eqn:pot2} and \ref{eqn:pot1} ) are computed using a global image representation, a $p$-dimensional image vector $g_i \in \mathcal{R}^{p}$, derived by the VGG convolutional neural network~\cite{vgg}.
Each potential value is computed by a linear regression with parameters, $\theta$, unique for each possible decision of verb and verb-role-noun (we refer to this as image regression in Figure~\ref{fig:model}), for example for the verb-role-noun potential in Equation \ref{eqn:pot1}:

\begin{equation}
\phi_e(v,e,n_e,i,\theta) = g_i^{T}\theta_{v,e,n_e}
\label{eqn:base_regression}
\end{equation}

Such a model does not directly represent the fact that nouns are reused between different roles, although the underlying neural network could hypothetically learn to encode such reuse during fine tuning. 
Instead, we introduce compositional potentials that make such reuse explicit. 

To formulate our compositional potential, we introduce a set of $m$-dimensional vectors $D = \{d_n \in \mathcal{R}^{m} | n \in N\}$, one vector for each noun in $N$, the set of nouns.
We create a set matrices $T = \{H_{(v,e)} \in \mathcal{R}^{p \times o} | (v,e) \in E_f\}$, one matrix for each verb, semantic role pair occurring in all frames $E_f$, that map image representations to $o$-dimensional verb-role representations.
Finally, we introduce a tensor of global composition weights, $C \in \mathcal{R}^{m \times o \times p}$.
We define a tensor weighting function, $T$, which takes as input a verb, $v$, semantic role, $e$, noun, $n$, and image representation, $g_i$ as:

\begin{equation}
T(v,e,n,g_i) = C \odot ( d_{n} \otimes g_i^{T}H_{(v,e)} \otimes g_i)
\label{eqn:tensor_feature}
\end{equation}

The tensor weighting function constructs an image specific verb-role representation by multiplying the global image vector and the verb-role matrix $g_i^{T}H_{(v,e)}$.  
Then, it combines a global noun representation, the image specific role representation, and the global image representation with outer products. 
Finally, it weights each dimension of the outer product with a weight from $C$. 
The weights in $C$ indicate which features of the 3-way outer product are important. 
The final potential is produced by summing up all of the elements of the tensor produced by $T$:

\begin{equation}
\phi_e(v,e,n_e,i) = \sum_{x=0}^{M}\sum_{y=0}^{O}\sum_{z=0}^{P} T(v,e,n_e,g_i)[x,y,z]
\label{eqn:tensor_sum}
\end{equation}

The tensor produced by $T$ in general will be high dimensional and very expressive.
This allows use of small dimensionality representations, making the function more robust to small numbers of samples for each noun.

The potential defined in Equation~\ref{eqn:tensor_sum} can be equivalently formulated as :
\begin{equation}
\phi_e(v,e,n_e,i) = g_{i}^{T}A(d_{n_e} \otimes g_{i}^{T}H_{(v,e)})
\label{eqn:tensor_sum_simplified}
\end{equation}
Where $A$ is a matrix with the same parameters as $C$ but flattened to layout the noun and role dimensions together. 
By aligning terms with Equation~\ref{eqn:base_regression}, one can see that tensor potential offers an alternative parametrized to the linear regression that uses many more general purpose parameters, those of $C$.
Furthermore, it eliminates any one parameter from ever being uniquely associated with one regression, instead compositionally using noun and verb-role representations to build up the parameters of the regression.
\subsection{Semantic Data Augmentation}
Situation recognition is strongly connected to language. 
Each situation can be thought of as simple declarative sentence about an activity happening in an image. 
For example, the first situation in Figure~\ref{fig:leading} could be expressed as ``man carrying baby on chest outside'' by knowing the prototypical ordering of semantic roles around verbs and inserting prepositions.
This relationship can be used to reduce semantic sparsity by using image search to find images that could contain the elements of a situations.

We convert annotated situations to phrases for semantic augmentation by exhaustively enumerating all possible sub-pieces of realized situations that occur in the imSitu training set (see Section~\ref{sec:setup} for implementation details). 
For example, in first situation of Figure~\ref{fig:leading}, we get the pieces: {\small $(\texttt{carrying},\{(\texttt{agent}, \texttt{man})\})$},
{\small$(\texttt{carrying},\{(\texttt{agent},\texttt{man}),(\texttt{item},\texttt{baby})\})$}, ect.
Each of these substructures is converted deterministically to a phrase using a template specific for every verb. 
For example, the template for carrying is ``\{\texttt{agent}\} carrying \{\texttt{item}\} \{with \texttt{agentpart}\} \{in \texttt{place}\}.''
Partial situations are realized into phrases by taking the first gloss in Wordnet of the synset associated with every noun in the substructure, inserting them into the corresponding slots of the template, and discarding unused slots.
For example, the phrases for the sub-pieces above are realized as ``man carrying'' and ``man carrying baby.'' 
These phrases are used to retrieve images from Google image search and construct a set, $W = \{(i,v,R_f)\}$, of images annotated with a verb and partially complete realized frames, by assigning retrieved images to the sub-piece that generated the retrieval query.\footnote{While these templates do not generate completely fluent phrases, preliminary experiments found them sufficiently accurate for image search because often no phrase could retrieve correct images. Longer phrases tended to have much lower precision.}

\paragraph{Pre-training}
Images retrieved from the web can be incorporated in a pre-training phase.
The images retrieved only have partially specified realized situations as labels. 
To account for this, we instead compute the marginal likelihood, $\hat{p}$, of the partially observed situations in $W$:
\begin{equation}
\begin{split}
\hat{p}(S |i; \theta) \propto \psi_v(v,i; \theta)\prod_{(e,n_e) \in R_f}\psi_e(v,e,n_e,i; \theta)\\
\times \prod_{e \notin R_f \wedge e \in E_f}\sum_{n}\psi_e(v,e,n,i; \theta)
\end{split}
\label{eqn:crf_pot}
\end{equation}
During pretraining, we optimize the marginal log-likelihood of $W$.
This objective provides a partial clustering over the unobserved roles left unlabeled during the retrieval process. 

\paragraph{Self Training}
Images retrieved from the web contain significant noise. 
This is especially true for role-noun combinations that occur infrequently, limiting their utility for pretraining. 
Therefore, we also consider filtering images in $W$ after a model has already been trained on fully supervised data from imSitu. 
We rank images in $W$ according to $\hat{p}$ as computed by the trained model and filter all those not in the top-$k$ for every unique $R_f$ in $W$.
We then pretrain on this subset of $W$, train again on imSitu, and then increase $k$. 
We repeat this process until the model no longer improves.
\section{Experimental Setup}
\label{sec:setup}


\paragraph{Models}
All models were implemented in Caffe~\cite{caffe} and use a pretrained VGG network~\cite{vgg} for the base image representation with the final two fully connected layers replaced with two fully connected layers of dimensionality 1024. We finetune all layers of VGG for all models.
For our tensor potential we use noun embedding size, $m = 32$, and role embedding size $o = 32$, and the final layer of our VGG network as the global image representation where $p = 1024$. Larger values of $m$ and $o$ did seem to improve results but were too slow to pretrain so we omit them.
In experiments where we use the image regression in conjunction with a compositional potential, we remove regression parameters associated with combinations seen fewer than 10 times on the imSitu training set to reduce overfitting.

\paragraph{Baseline}
We compare our models to two alternative methods for introducing effective sharing between nouns.
The first baseline (Noun potential in Table \ref{tab:full_results_dev} and \ref{tab:rare_results_dev}) adds a potential into the baseline CRF for nouns independent of roles.
We modify the probability, from Equation~\ref{eqn:crf_pot} of a situation, $S$, given an image $i$, to not only decompose by pairs of roles, $e$ and nouns $n_e$ in a realized frame $R_f$, but also nouns $n_e$:   
\begin{equation}
p(S |i; \theta) \propto \psi_v(v,i; \theta)\prod_{(e,n_e) \in R_f} \psi_e(v,e,n_e,i; \theta)\psi_{n_e}(n_e,i)
\label{eqn:crf_pot2}
\end{equation}
The added potential, $\psi_{n_e}$, is computed using a regression from a global image representation for each unique $n_e$.  

The second baseline we consider is compositional but does not use a tensor based composition method.
The model instead constructs many verb-role representations and combines them with noun representations using inner-products (Inner product composition in Table \ref{tab:full_results_dev} and \ref{tab:rare_results_dev}). 
In this model, as in the tensor model in Section~\ref{sec:model}, we use a global image representation $g_i \in \mathcal{R}^p$ and a set noun vectors, $d_n \in \mathcal{R}^m$ for every noun $n$.  
We also assume $t$ verb-role matrices $H_{t,v,e} \in \mathcal{R}^{o \times p}$ for every verb-role in $E_f$.
We compute the corresponding potential as in Equation~\ref{eqn:tensor_sum_simplified}:
\begin{equation}
\phi_e(v,e,n_e,i) = \sum_{k} d_{n_e}^{T}H_{(k,v,e)}q_i
\label{eqn:tensor_sum_simplified}
\end{equation}
The model is motivated by compositional models used for semantic role labeling~\cite{fitz_srl} and allows us to trade-off the need to reduce parameters associated with nouns and expressivity. 
We grid search values of $t$ such that $t \cdot o $ was at most 256, the largest size network we could afford to run and $o = m$, a requirement on the inner product. We found the best setting at $t = 16$, $o = m = 16$. 

\paragraph{Decoding}
We experimented with two decoding methods for finding the best scoring situation under the CRF models. 
Systems which used the compositional potentials performed better when first predicting a verb $v^m$ using the max-marginal over semantic roles:
$v^{m} = \argmax_v\sum_{(e,n_e)} p(v, R_f | i)$
and then predict a realized frame, $R_f^{m}$, with max score for $v^m$:
$R_f^{m} = \argmax_{R_f} p(v^{m}, R_f | i ) $.
All other systems performed better maximizing jointly for both verb and realized frame.

\begin{table*}
\footnotesize
\begin{center}
\begin{tabular}{ |c | c | c | c | c | c || c | c | c || c | c || c | }
\hline
&&& \multicolumn{3}{ |c|| }{top-1 predicted verb} & \multicolumn{3}{ |c|| }{top-5 predicted verbs} & \multicolumn{2}{ |c|| }{ground truth verbs} & \multicolumn{1}{ |c| }{} \\
 &&& verb	 & value	 & value-all   & verb	 & value	 & value-all  & value	 & value-all & mean \\
 \hline
\parbox[t]{2mm}{\multirow{5}{*}{\rotatebox[origin=c]{90}{imSitu}}}
& 1 & Baseline: Image Regression \cite{yatskar2016} & 32.25  & 24.56 & 14.28 & 58.64 & 42.68 & 22.75  & 65.90 & 29.50 & 36.32 \\
\cline{2-11}
& 2 & Noun Potential + reg & 27.64  & 21.21 & 12.21 & 53.95 & 39.95 & 21.45  & 68.87 & 32.31 & 34.70 \\
\cline{2-11}
& 3 & Inner product composition + reg & 32.13  & 24.77 & 14.71 & 58.33 & 42.93 &  23.14 & 66.79 & 30.2 & 36.62  \\  
\cline{2-11}
& 4 & Tensor composition & 31.73 & 24.04  & 13.73 & 58.06 & 42.64 & 22.7 & 68.73 & 32.14 & 36.72  \\  
\cline{2-11}
& 5 & Tensor composition + reg & {32.91}  & {25.39} & {14.87} & {59.92}  & {44.5}  & {24.04} & {69.39} & {33.17} & {38.02} \\   
\hline
\hline
\parbox[t]{2mm}{\multirow{3}{*}{\rotatebox[origin=c]{90}{ + SA }}}
& 6 & Baseline : Image Regression & 32.40  & 24.14 & 15.17 & 59.10 & 44.04  & 24.40 & 68.03 & 31.93 & 37.53\\
\cline{2-11}
& 7 & Tensor composition + reg  & 34.04  & 26.47  & \textbf{15.73} & 61.75 & 46.48  & \textbf{25.77} & \textbf{70.89} & \textbf{35.08} & 39.53 \\
\cline{2-11}
& 8 & Tensor composition + reg + self train & \textbf{34.20} & \textbf{26.56} & 15.61 & \textbf{62.21}  & \textbf{46.72}  & 25.66 & 70.80 & 34.82 & \textbf{39.57} \\ 
\hline
\end{tabular}
\caption{\small Situation recognition results on the full imSitu development set. The results are divided by models which were only trained on imSitu data, rows 1-5, and models which use web data through semantic data augmentation, marked as +SA in rows 6-8. Models marked with +reg also include image regression potentials used in the baseline. Our tensor composition model, row 5, significantly outperforms the existing state of the art, row 1, addition of a noun potential, row 2, and a compositional baseline, row 3. The tensor composition model is able to make better use of semantic data augmentation (row 8) than the baseline (row 6). }
\label{tab:full_results_dev}
\end{center}
\vspace{-20pt}
\end{table*}

\begin{table*}
\footnotesize
\begin{center}
\begin{tabular}{ | c |c | c | c | c | c || c | c | c || c | c || c | }
\hline
&&&  \multicolumn{3}{ |c|| }{top-1 predicted verb} & \multicolumn{3}{ |c|| }{top-5 predicted verbs} & \multicolumn{2}{ |c|| }{ground truth verbs} & \multicolumn{1}{ |c| }{} \\
 &&& verb	 & value	 & value-all   & verb	 & value	 & value-all  & value	 & value-all & mean \\
 \hline
\parbox[t]{2mm}{\multirow{4}{*}{\rotatebox[origin=c]{90}{imSitu}}}
& 1 &Baseline: image regression \cite{yatskar2016} & 19.89  & 11.68 & \textbf{2.85} & 44.00 & 24.93  & \textbf{6.16}  & 50.80 & 9.97 & 19.92 \\
\cline{2-11}
& 2 & Noun potential + reg & 15.88  & 9.13 & 1.86 & 38.22 & 22.28  & 5.46  & 54.65 & 11.91 & 19.92 \\
\cline{2-11}
& 3 &Inner product composition + reg & 18.96  & 10.69& 1.89 & 42.53 & 23.28  & 3.69  & 49.54 & 6.46 & 19.63 \\
\cline{2-11}
& 4 &Tensor composition& 19.78   & 11.28 & 2.26 & 42.66  & 24.42  & 5.57 & 54.06 & 11.47 & 21.43  \\ 
\cline{2-11}
& 5 &Tensor composition + reg & \textbf{21.12}  & 11.89 & 2.20 & 45.14 & 25.51  & 5.36 & 53.58 & 10.62 & 21.93  \\    
\hline
\hline
\parbox[t]{2mm}{\multirow{3}{*}{\rotatebox[origin=c]{90}{ + SA }}}
& 6 &Baseline : image regression & 19.95  & 11.44 & 2.13 & 43.08 & 24.56  & 4.95 & 51.55 & 8.41 & 20.76\\
\cline{2-11}
& 7 & Tensor composition + reg  & 20.08  & 11.58  & 2.22  & 44.82 & 26.02 & 5.55 & 55.45 & 11.53 & 22.16 \\
\cline{2-11}
& 8 &Tensor composition + reg + self train & 20.52 & \textbf{11.91}  & 2.34 & \textbf{45.94} & \textbf{26.99}  & 6.06 & \textbf{55.90} & \textbf{12.04} & \textbf{22.71} \\ 
\hline
\end{tabular}
\caption{\small Situation prediction results on the rare portion imSitu development set. The results are divided by models which were only trained on imSitu data, rows 1-5, and models which use web data through semantic data augmentation, marked as +SA in rows 6-8. Models marked with +reg also include image regression potentials used in the baseline. Semantic data augmentation with the baseline hurts for rare cases.  Semantic augmentation yields larger relative improvement on rare cases and a composition-based model is required to realize these gains. }
\label{tab:rare_results_dev}
\end{center}
\vspace{-20pt}
\end{table*}

\paragraph{Optimization}
All models were trained with stochastic gradient descent with momentum 0.9 and weight decay 5e-4. 
Pretraining in semantic augmentation was conducted with initial learning rate of 1e-3, gradient clipping at 100, and batch size 360. 
When training on imSitu data, we use an initial learning rate of 1e-5. 
For all models, the learning rate was reduced by a factor of 10 when the model did not improve on the imSitu dev set.

\paragraph{Semantic Augmentation}
In experiments with semantic augmentation, images were retrieved using Google image search. 
We retrieved 200 medium sized, full-color, safe search filtered images per query phrase.
We produced over 1.5 million possible query phrases from the imSitu training set, the majority extremely rare.
We limited the phrases to any that occur between 10 and 100 times in imSitu and for phrases that occur between 3 and 10 times we accepted only those containing at most one noun.
Roughly 40k phrases were used to retrieve 5 million images from the web.
All duplicate images occurring in imSitu were removed.
For pretraining, we ran all experiments up to 50k updates (roughly 4 epochs). 
For self training, we only self train on rare realized frames (those 10 or fewer times in imSitu train set). 
Self training yielded diminishing gains after two iterations and we ran the first iteration at k=10 and the second at k=20.

\paragraph{Evaluation} 
We use the standard data split for imSitu\cite{yatskar2016} with 75k train, 25k development, and 25k test images.
We follow the evaluation setup defined for imSitu, evaluating verb predictions (verb) and semantic role-value pair predictions (value) and full structure correctness (value-all). 
We report accuracy at top-1, top-5 and given the ground truth verb and the average across all measures (mean). 
We also report performance for examples requiring rare (10 or fewer examples in the imSitu training set) predictions. 

\section{Results}
\label{sec:results}


\begin{table*}
\footnotesize
\begin{center}
\begin{tabular}{ | c |c | c | c | c | c || c | c | c || c | c || c | }
\hline
\multicolumn{2}{|c|}{}&&  \multicolumn{3}{ |c|| }{top-1 predicted verb} & \multicolumn{3}{ |c|| }{top-5 predicted verbs} & \multicolumn{2}{ |c|| }{ground truth verbs} & \multicolumn{1}{ |c| }{} \\
\multicolumn{2}{|c|}{}&& verb	 & value	 & value-all   & verb	 & value	 & value-all  & value	 & value-all & mean \\
 \hline
\multicolumn{2}{|c|}{\multirow{2}{*}{\rotatebox[origin=c]{}{ imSitu }}}
&Baseline: Image Regression \cite{yatskar2016} & 32.34  & 24.64 & 14.19 & 58.88 & 42.76  & 22.55 & 65.66 & 28.96 & 36.25 \\
\cline{3-11}
\multicolumn{2}{|c|}{}&Tensor composition + reg & 32.96 & 25.32 & 14.57 & 60.12  & 44.64  & 24.00 & 69.2 & 32.97 & 37.97  \\    
\hline
\hline
\multicolumn{2}{|c|}{\multirow{2}{*}{\rotatebox[origin=c]{}{ + SA }}}
&Baseline : Image Regression & 32.3  & 24.95  & 14.77 & 59.52 & 44.08  & 23.99 & 67.82 & 31.46 & 37.36\\
\cline{3-11}
\multicolumn{2}{|c|}{}&Tensor composition + reg + self train & \textbf{34.12}  & \textbf{26.45}  & \textbf{15.51} & \textbf{62.59} & \textbf{46.88} & \textbf{25.46} & \textbf{70.44} & \textbf{34.38} & \textbf{39.48} \\ 
\hline
\end{tabular}
\caption{\small Situation prediction results on the full imSitu test set. Models were run exactly once on the test set. General trends are identical to experiments run on development set.}
\label{tab:full_results_test}
\end{center}
\vspace{-20pt}
\end{table*}

\begin{table*}
\footnotesize
\begin{center}
\begin{tabular}{ | c |c | c | c | c | c || c | c | c || c | c || c | }
\hline
\multicolumn{2}{|c|}{\multirow{2}{*}{}}&&  \multicolumn{3}{ |c|| }{top-1 predicted verb} & \multicolumn{3}{ |c|| }{top-5 predicted verbs} & \multicolumn{2}{ |c|| }{ground truth verbs} & \multicolumn{1}{ |c| }{} \\
\multicolumn{2}{|c|}{}&& verb	 & value	 & value-all   & verb	 & value	 & value-all  & value	 & value-all & mean \\
 \hline
\multicolumn{2}{|c|}{\multirow{2}{*}{\rotatebox[origin=c]{}{ imSitu }}}
&Baseline: Image Regression \cite{yatskar2016} & \textbf{20.61}  & 11.79 & \textbf{3.07}  & 44.75 & 24.85 & 5.98 & 50.37 & 9.31 & 21.34 \\
\cline{3-11}
\multicolumn{2}{|c|}{}&Tensor composition + reg & 19.96  & 11.57& 2.30 & 44.89 & 25.26  & 4.87 & 53.39 & 10.15 & 21.55  \\    
\hline
\hline
\multicolumn{2}{|c|}{\multirow{2}{*}{\rotatebox[origin=c]{}{ + SA }}}
&Baseline : Image Regression & 19.46  & 11.15& 2.13 & 43.52 & 24.14  & 4.65 & 51.21 & 8.26 & 20.57\\
\cline{3-11}
\multicolumn{2}{|c|}{}&Tensor composition + reg + self train & 20.32  & \textbf{11.87}  & 2.52 & \textbf{47.07} & \textbf{27.50}  & \textbf{6.35} & \textbf{55.72} & \textbf{12.28} & \textbf{22.95} \\ 
\hline
\end{tabular}
\caption{\small Situation prediction results on the rare portion of imSitu test set. Models were run exactly once on the test set. General trends established on the development set are supported. }
\label{tab:rare_results_test}
\end{center}
\vspace{-20pt}
\end{table*}

\paragraph{Compositional Tensor Potential}
Our results on the full imSitu dev set are presented in Table~\ref{tab:full_results_dev} in rows 1-5.   
Overall results demonstrate that adding a noun potential (row 2) and our baseline composition model (row 3) are ineffective and perform worse than the baseline CRF (row 1).
We hypothesize that systematic variation in object appearance between roles is challenging for these models. 
Our tensor composition model (row 4) is able to better capture such variation and effectively share information among nouns, reflected by improvements in value and value-all accuracy given ground truth verbs while maintaining high top-1 and top-5 verb accuracy.
However, as expected, many situations cannot be predicted only compositionally based on nouns (consider that a horse sleeping looks very different than a horse swimming and nothing like a person sleeping).
Combination of the image regression potential and our tensor composition potential (row 5) yields the best performance, indicating they are modeling complementary aspects of the problem.
Our final model (row 5) only trained on imSitu data outperforms the baseline on every measure, improving over 1.70 points overall. 

Results on the rare portion of the imSitu dataset are presented in Table~\ref{tab:rare_results_dev} in rows 1-5.
Our final model (row 5) provides the best overall performance (mean column) on rare cases among models trained only on imSitu data, improving by 0.64 points on average.
All models struggle to get correctly entire structures (value-all columns), indicating rare predictions are extremely hard to get completely correct while the baseline model which only uses image regression potentials performs the best. 
We hypothesize that image regression potentials may allow the model to more easily coordinate predictions across roles simultaneously because role-noun combinations that always co-occur will always have the same set of regression weights.


\paragraph{Semantic Data Augmentation}
Our results on the full imSitu development set are presented in Table~\ref{tab:full_results_dev} in rows 6-8. 
Overall results indicate that semantic data augmentation helps all models, while our tensor model (row 7) benefits more than the baseline (row 6).
Self training improves the tensor model slightly (row 8), making it perform better on top-1 and top-5 predictions but hurting performance given gold verbs.
On average, our final model outperforms the baseline CRF trained on identical data by 2.04 points. 

Results on the rare portion of the imSitu dataset are presented in Table~\ref{tab:rare_results_dev} in rows 6-8. 
Surprisingly, on rare cases semantic augmentation hurts the baseline CRF (line 6). 
Rare instance image search results are extremely noisy. 
On close inspection, many of the returned results do not contain the target activity at all but instead contain target nouns. 
We hypothesize that without an effective global noun representation, the baseline CRF cannot extract meaningful information from such extra data. 
On the other hand, our tensor model (line 7) improves on these rare cases overall and with self training improves further (line 8).

\begin{figure}
\centering
\hspace*{-10pt}
\includegraphics[width=0.5\textwidth]{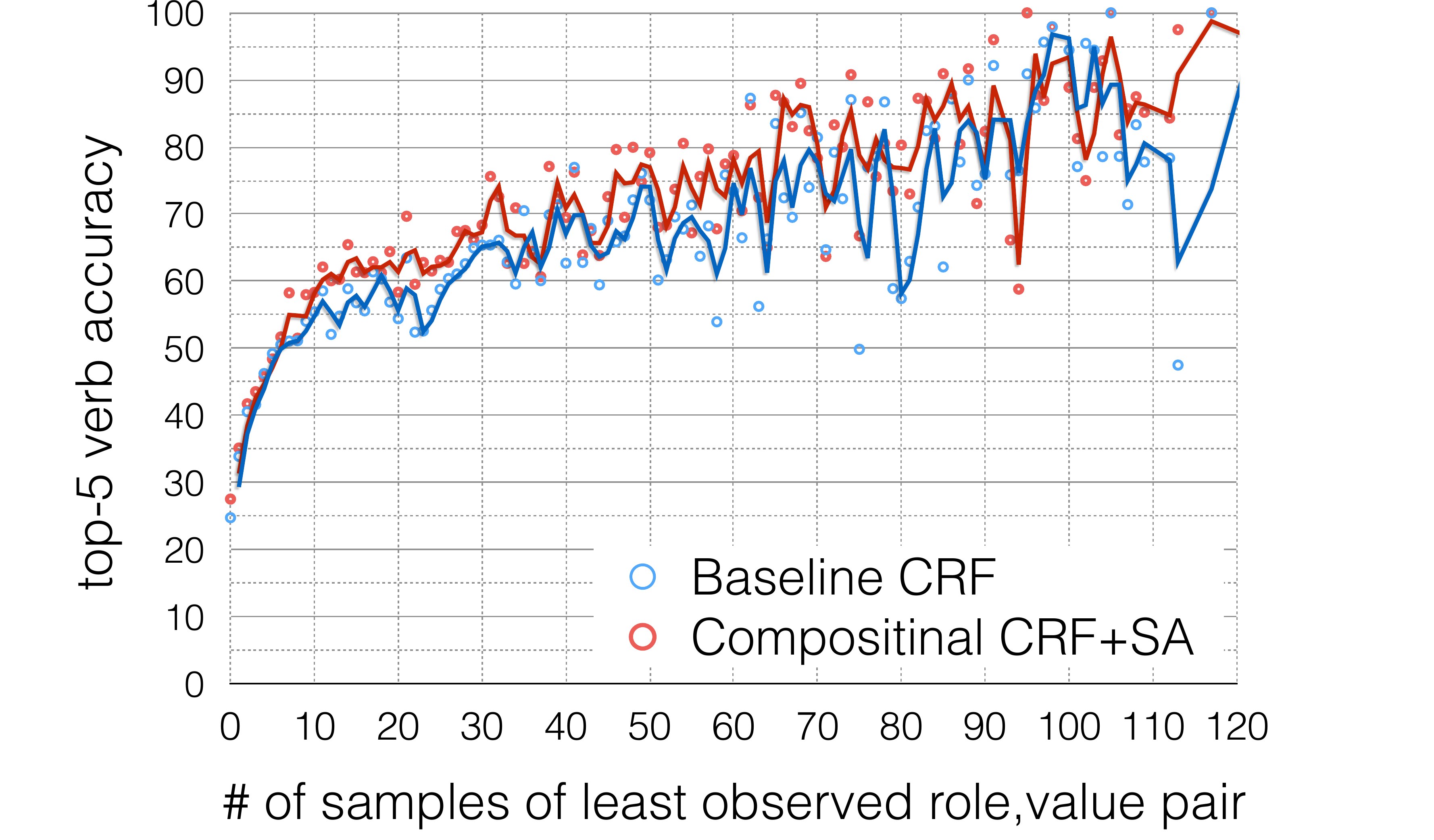}
\caption{\small Top-5 verb accuracy on the imSitu development set. Our final compositional CRF with semantic data augmentation outperforms the baseline CRF on rare cases (fewer than 10 training examples), but both models continue to struggle with semantic sparsity. For our final model, the largest improvement relative to the baseline are for cases with 5-35 examples on the training set.}
\label{fig:summarize_results}
\end{figure}

\begin{figure*}[]
\centering
\includegraphics[width=.95\textwidth]{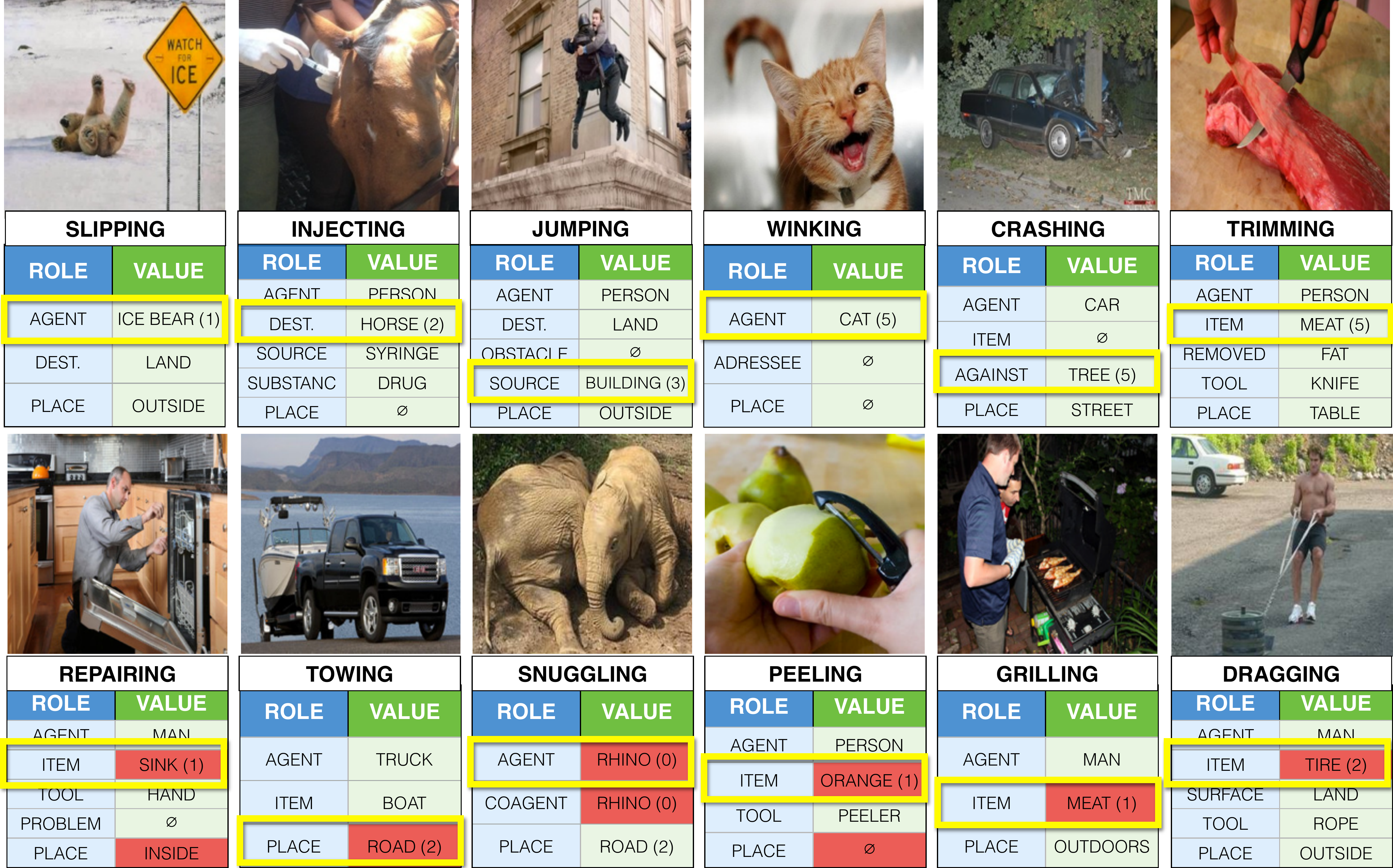}
\caption{\small Output from our final model on development examples containing rare role-noun pairs. The first row contains examples where the model correctly predicts the entire structures in the top-5 (top-5, value-all). We highlight the particular role-noun pairs that make the examples rare with a yellow box and put in number occurances of it in the imSitu training set. The second row contains examples where the verb was correctly predicted in the top-5 but not all the values were predicted correctly. We highlight incorrect predictions in red. Many such predictions occurr zero times in the training set (ex. the third image on the second row). All systems struggle with such cases.  
}
\label{fig:qual}
\end{figure*}

\paragraph{Overall Results}
Experiments show that (a) our tensor model is able perform better in comparable data settings, (b) our semantic augmentation techniques largely benefit all models, and (c) our tensor model benefits more from semantic augmentation. 
We also present our full performance on top-5 verb across all numbers of samples in Figure~\ref{fig:summarize_results}. 
While our compositional CRF with semantic augmentation outperforms the baseline CRF, both models continue to struggle on uncommon cases. 
Our techniques seem to give most benefit for examples requiring predictions of structures seen between 5 and 35 times, while providing somewhat less benefit to even rarer ones. 
It is challenging future work to make further improvements for extremely rare outputs. 

We also evaluated our models on the imSitu test set exactly once. 
The results are summarized in Table~\ref{tab:full_results_test} for the full imSitu test set and in Table~\ref{tab:rare_results_test} for the rare portion. 
General trends established on the imSitu dev set are supported.
We provide examples in Figure~\ref{fig:qual} of predictions our final system made on rare examples from the development set.


\section{Related Work}





Learning to cope with semantic sparsity is closely related to zero-shot or k-shot learning. 
Attribute-based learning \cite{lampert2014attribute,lampert,attributes}, cross-modal transfer \cite{lapata, lei2015predicting, devise, wampimuk} and using text priors \cite{lu2016visual,youtube2txt} have all been proposed but they study classification or other simplified settings. 
For the structured case, image captioning models~\cite{me,stanfordcaption,larrycaption,msrcaption,baidu,hodosh,flikr1m,coco} have been observed to suffer from a lack of diversity and generalization~\cite{googlecaption}.
Recent efforts to gain insight on such issues extract subject-verb-object (SVO) triplets from captions and count prediction failures on rare tuples \cite{atzmon2016learning}. 
Our use of imSitu to study semantic sparsity circumvents the need for intermediate processing of captions and generalizes to verbs with more than two arguments.

Compositional models have been explored in a number of applications in natural language processing, such as sentiment analysis~\cite{richard_sentiment}, dependency parsing~\cite{lei2014low},  text similarity~\cite{baroni_tensor}, and visual question answering~\cite{jacob_vqa} as effective tools for combining natural language elements for prediction. 
Recently, bilinear pooling~\cite{bilinearpool} and compact bilinear pooling~\cite{compactbilinear} have been proposed as second-order feature representations for tasks such as fine grained recognition and visual question answer. 
We build on such methods, using low dimensional embeddings of semantic units and expressive outer product computations. 

Using the web as a resource for image understanding has been studied through NEIL~\cite{chen2013neil}, a system which continuously queries for concepts discovered in text, and Levan~\cite{levan}, which can create detectors from user specified queries. 
Web supervision has also been explored for pretraining convolutional neural networks~\cite{chen2015webly} or for fine-grained bird classification~\cite{chen2015webly} and common sense reasoning~\cite{viske}.
Yet we are the first to explore the connection between semantic sparsity and language for automatically generating queries for semantic web augmentation and we are able to show improvement on a large scale, fully supervised structured prediction task. 

\section{Conclusion}
We studied situation recognition, a prototypical instance of a structured classification problem with significant semantic sparsity. 
Despite the fact that the vast majority of the possible output configurations are rarely observed in the training data, we showed it was possible in introduce new compositional models that effectively share examples among required outputs and semantic data augmentation  techniques  that  significantly improved performance. 
In the future, it will be important to introduce similar techniques for related problems with semantic sparsity and generalize these ideas to the zero-shot learning. 

{\small
\bibliographystyle{ieee}
\bibliography{visual}
}

\end{document}